\documentclass[conference]{IEEEtran}
\IEEEoverridecommandlockouts
\usepackage{cite}
\usepackage[numbers,sort&compress]{natbib}

\usepackage[utf8]{inputenc}
\usepackage{amsmath,amssymb}
\usepackage{graphicx}
\usepackage{bm}
\usepackage{enumerate}
\usepackage{comment}
\usepackage{xcolor}
\usepackage{url}
\usepackage{color,soul}
\usepackage{subcaption}
\usepackage{float}

\definecolor{light-gray}{gray}{0.8}

\usepackage{amsmath,amssymb,amsfonts}
\usepackage{algorithmic}
\usepackage{graphicx}
\usepackage{textcomp}
\usepackage{xcolor}
\usepackage{subcaption}

\def\BibTeX{{\rm B\kern-.05em{\sc i\kern-.025em b}\kern-.08em
    T\kern-.1667em\lower.7ex\hbox{E}\kern-.125emX}}

\makeatletter
\newcommand{\linebreakand}{%
  \end{@IEEEauthorhalign}
  \hfill\mbox{}\par
  \mbox{}\hfill\begin{@IEEEauthorhalign}
}
\makeatother

\begin{document}

\title{An Intelligent Fault Self-Healing Mechanism for Cloud AI Systems via Integration of Large Language Models and Deep Reinforcement Learning\\}

\author{

\small 

\begin{tabular}[t]{c@{\extracolsep{8em}}c} 

1\textsuperscript{st} Ze Yang \textsuperscript{*}  & 2\textsuperscript{nd} Yihong Jin \\
\textit{University of Illinois Urbana-Champaign} & \textit{University of Illinois Urbana-Champaign} \\
Champaign, USA & Champaign, USA \\
\textsuperscript{*}Corresponding Author: zeyang2@illinois.edu & yihongj3@illinois.edu\\

\\

3\textsuperscript{rd} Juntian Liu & 4\textsuperscript{th} Xinhe Xu \\
\textit{Computer Science Department} & \textit{Computer Science Department} \\
\textit{University of Illinois Urbana-Champaign} & \textit{University of Illinois Urbana-Champaign} \\
Champaign, USA & Champaign, USA \\
jl203@illinois.edu & xinhexu2@illinois.edu  \\

\\
\end{tabular}
}

\maketitle

\begin{abstract}
As the scale and complexity of cloud-based AI systems continue to increase, the detection and adaptive recovery of system faults have become the core challenges to ensure service reliability and continuity. In this paper, we propose an Intelligent Fault Self-Healing Mechanism (IFSHM) that integrates Large Language Model (LLM) and Deep Reinforcement Learning (DRL), aiming to realize a fault recovery framework with semantic understanding and policy optimization capabilities in cloud AI systems. On the basis of the traditional DRL-based control model, the proposed method constructs a two-stage hybrid architecture: (1) an LLM-driven fault semantic interpretation module, which can dynamically extract deep contextual semantics from multi-source logs and system indicators to accurately identify potential fault modes; (2) DRL recovery strategy optimizer, based on reinforcement learning, learns the dynamic matching of fault types and response behaviors in the cloud environment. The innovation of this method lies in the introduction of LLM for environment modeling and action space abstraction, which greatly improves the exploration efficiency and generalization ability of reinforcement learning. At the same time, a memory-guided meta-controller is introduced, combined with reinforcement learning playback and LLM prompt fine-tuning strategy, to achieve continuous adaptation to new failure modes and avoid catastrophic forgetting. Experimental results on the cloud fault injection platform show that compared with the existing DRL and rule methods, the IFSHM framework shortens the system recovery time by $37\%$ with unknown fault scenarios. 
\end{abstract}


\section{Introduction}
With the widespread deployment of cloud computing, container orchestration, microservice architectures, and AI inference services in industry and academia, modern cloud-based AI systems are becoming the core support of the infrastructure of an intelligent society \cite{li2024advances}. These systems not only need to support high-throughput concurrent request processing, but also meet multiple requirements such as dynamic scaling, service elasticity, high availability, and IntelliSense. Under such multi-dimensional constraints, the architecture of the system is becoming increasingly complex, and it usually contains multiple heterogeneous subsystems and service modules, such as scheduling managers, inference accelerators, container runtimes, load balancers, asynchronous message queues, distributed caches, deep learning inference engines, and etc \cite{yang2025research}. More importantly, these subsystems often operate together in a "weak coupling and strong dependency" manner, and even if a single component undergoes a slight performance degradation, it may trigger cascading failures due to the dependency chain amplification effect, which will eventually lead to the overall performance degradation of the system or even the unavailability of services \cite{rana2025ai}.  Modeling and understanding these complex dependencies, often represented as graphs, is a significant challenge in itself \cite{2023arXiv231003272H, liu2024graphsnapshot}. This fault propagation is highly nonlinear and uncertain, which seriously threatens the stability and service continuity of cloud AI systems.

In this context, timely detection, diagnosis, and recovery of system faults have become an important means to ensure system reliability and reduce operation and maintenance costs \cite{ding2024confidence}. However, modern cloud system failure types are not only numerous and complex, but also dynamically evolving, semantically ambiguous, and even deceptive \cite{li2024deception, 10628639}. For example, a hardware failure may manifest itself as a "continuous spike in CPU usage", while a software failure may manifest itself as a "thread blocking" or "dependent service interface unresponsive" \cite{alonso2021optimization}. To further complicate matters, many fault signals are cross-modal, multi-sourced, and unstructured, making accurate extraction of fault context a huge challenge. Various studies have been done t

For a long time, the fault detection and recovery mechanism mainly relied on manual empirical rules and static scripts. This type of system relies on rules defined by O\&M experts, such as keyword matching, threshold triggering, and abnormal indicator alarms, to identify faults, and then trigger fixed recovery actions (such as restarting services, releasing memory, etc.) \cite{yang2024hades}. Although it can be applied in small-scale systems, it lacks generalization ability, rigid strategy, and cannot adapt to the dynamic evolution of the system, which has gradually exposed serious limitations. 

In order to solve the above problems, some studies have begun to try methods based on system modeling, such as using Bayesian networks, Petri networks, Markov chains, etc. to describe the relationship between component state transitions, and predicting possible system anomalies based on model analysis. However, such methods often rely on accurate modeling of the internal operating mechanisms of the system, which is often difficult to implement in actual production systems \cite{jin2025adaptive}. Especially in multi-cloud, hybrid deployment environments, modeling errors caused by service migration, configuration drift, or platform heterogeneity can lead to model prediction failures. Large Language Models (LLMs) and generative models could potentially be helpful in this circumstance, as they can learn patterns from data and adapt to changing system dynamics, providing a more flexible and robust approach to anomaly detection and prediction \cite{li2024exploring, li2024contextual, yang2024comparative, beharidecision, sehanobishscalable, baral2024automated, ji2025cloud, he2025givestructuredreasoninglarge, yu2024large}.

\section{RELATED WORK}
Karamthulla et al. \cite{karamthulla2023ai} analyzed the application of AI techniques such as machine learning and neural networks in creating self-healing mechanisms through a series of case studies, and discussed challenges such as scalability, adaptability, and robustness. The findings contribute to advancing understanding of the role of AI in enhancing the fault tolerance and resilience of engineering platforms Vemula \cite{vemula2025ai} explores an AI-enhanced, self-healing cloud architecture that leverages advanced AI algorithms to automatically detect, diagnose, and remediate anomalies in real-time. By introducing self-healing mechanisms, these architectures ensure data integrity and privacy while maintaining operational efficiency. 

Furthermore, Nama et al. \cite{nama2024artificial} begin with a comprehensive review of current automated testing practices and the limitations of existing testing systems, and then demonstrate the role of AI in software testing, specifically how machine learning can enhance the accuracy of fault detection and prediction. Through predictive maintenance strategies and data analysis methods, AI is being explored for early detection and prevention of failures. Feng et al. \cite{feng2025integration} focused on their potential in creating fully autonomous self-healing control architectures by combining multi-agent systems (MASs) with advanced artificial intelligence (AI) algorithms. Combining MASs with IEC 61850 communication standards, a fault diagnosis, isolation and recovery mechanism has been developed specifically for metro systems.

Pentyala \cite{pentyala2024artificial} proposes an AI-based framework that leverages machine learning (ML) and deep learning (DL) models to analyze system logs, metrics, and telemetry data to identify anomalies in real time, predict potential failures, and perform automated remediation before failures occur. Li et al. \cite{li2021automated} proposed an automated intelligent self-healing system (AIHS) for cloud-scale data centers. AIHS combines machine learning technology to achieve scalable self-healing capabilities for cloud data centers. The system significantly improves the reliability and availability of the data center through real-time monitoring, fault detection, and automatic repair.

\section{METHODOLOGIES}
\subsection{Fault semantic interpretation and recovery policy}
The fault performance of cloud systems is heterogeneous, multimodal, and semantic dependent. For example, ``The CPU is abnormally high'', ``The log indicates that the GC failed'', ``Memory overflow'', and ``Response timeout'' may indicate the same root cause. Traditional methods are difficult to capture this kind of semantic mapping relationship, while LLM has strong context modeling and cross-modal embedding capabilities. Let's start by constructing a uniform representation of the input, as shown in Equation 1:

\begin{equation}
X_t = \left\{ x_t^L, x_t^M, x_t^A \right\}, \tag{1}
\end{equation}

where $x_t^L \in \mathbb{R}^{n_L}$ is a structured or unstructured log fragment; $x_t^M \in \mathbb{R}^{n_M}$ is the time window for system resource metrics, such as CPU, memory, and disk, which can be effectively modeled using advanced time-series analysis techniques\cite{ni2024timeseries}. $x_t^A \in \mathbb{R}^{n_A}$ is a multi-dimensional alarm output by the system exception or abnormal inference module. We map the multimodal input to the unified context space via a modal-aware encoder $f_{\text{enc}}$, as shown in Equation 2:

\begin{equation}
h_t = f_{\text{enc}}(X_t) = \text{Concat} \left( E_L(x_t^L), E_M(x_t^M), E_A(x_t^A) \right). \tag{2}
\end{equation}

Then input the LLM to obtain the final semantic state vector through deep attention computation \cite{ji-etal-2024-rag}, as shown in Equation 3:

\begin{equation}
z_t = L_\theta(h_t). \tag{3}
\end{equation}

The attention mechanism allows the model to dynamically focus on the information positions that are strongly related to the fault characteristics, such as key abnormal words (``timeout'', ``OOM'') or the peak points of key indicators, so as to enhance the causality and discriminant nature of state representation. In order to efficiently represent complex recovery operations (e.g., ``cold migrate a container'', ``release node cache'', ``reload service configuration''), we introduce hierarchical action space modeling, as shown in Equation 4:

\begin{equation}
a_t = \left< a_t^{(h)}, a_t^{(l)} \right>, \tag{4}
\end{equation}

Among them, high-level action $a_t^{(h)}$ is the policy category selection, which has a semantic prior (such as restart, migration, and elastic scale-in). Low-level actions $a_t^{(l)}$ are parameterized execution details, such as target node ID, resource ratio, delay policy, etc. We design a hierarchical strategy network as shown in Equation 5:

\begin{equation}
\pi(s_t) = \pi_h(s_t) \cdot \pi_l(a_t^{(h)}, s_t). \tag{5}
\end{equation}

Reinforcement learning uses a distributed PPO algorithm, and its core objective function includes a strategy clamping mechanism and advantage estimation, as shown in Equations 6 and 7:

\begin{equation}
L_{\text{PPO}} = \mathbb{E}_t \left[ r_t(\phi)\hat{A}_t, \text{clip}(r_t(\phi), 1 - \epsilon, 1 + \epsilon)\hat{A}_t \right] \tag{6}
\end{equation}

\begin{equation}
r_t(\phi) = \frac{\pi_\phi(s_t)}{\pi_{\phi_{\text{old}}}(s_t)}. \tag{7}
\end{equation}

where $\hat{A}_t$ is the generalized advantage estimation, which is used for more stable advantage calculations.

\subsection{Memory-guided meta-control}
In addition, we introduce information bottleneck regular terms, such as Equation 8:

\begin{equation}
L_{IB} = \beta \cdot D_{KL}\left[\pi(s) \parallel q(a)\right]. \tag{8}
\end{equation}

It is used to control the complexity of the strategy, avoid overfitting the strategy, and improve the generalization ability.

Catastrophic forgetting is a common problem during DRL training, especially when the system is faced with ``non-stationary task shifting'' such as new types of failures. To this end, we introduce a memory-enhancing meta-controller $M_w$ to perform cross-task memory management and cue optimization, enabling a form of associative reasoning even from limited prior examples of faults\cite{he2025selfgiveassociativethinkinglimited}. Record high-value trajectories, as shown in Equation 9:

\begin{equation}
B = \left\{ \left(s_i, a_i, r_i, s_i'\right) \right\}. \tag{9}
\end{equation}

Define the sampling probability based on the TD error, as shown in Equation 10:

\begin{equation}
P_i \propto |\delta_i| + \epsilon. \tag{10}
\end{equation}

Low-frequency but critical sample reuse increases the frequency of empirical playback of rare fault types, a strategy conceptually linked to curriculum-guided reinforcement learning which also aims to optimize the learning process\cite{ji2025curriculum}. When the system finds an unknown failure mode, it triggers a prompt learning adjustment for the LLM, as shown in Equation 11:

\begin{equation}
z_t^{new} = L_\theta\left(\text{Prompt}(h_t; w)\right). \tag{11}
\end{equation}

where \texttt{Prompt} is a ``soft template'' learned through gradient descent, and the LLM is guided to focus on specific semantic patterns as follows: CPU/Memory Relief and Dependent Service Exception Failures (API Request Failures).

The goal of optimization is to minimize the semantic differences in reconstruction, as shown in Equation 12:

\begin{equation}
L_{\text{Prompt}} = \left\| z_t^{\text{new}} - z_t^{\text{target}} \right\|_2^2 \tag{12}
\end{equation}

Ultimately, the training objectives of the IFSHM system consist of the following multi-task losses, as in Equation 13:

\begin{equation}
L_{\text{total}} = L_{PPO} + \lambda_1 L_V + \lambda_2 L_{IB} + \lambda_3 L_{\text{Prompt}} + \lambda_4 L_{\text{Cluster}}. \tag{13}
\end{equation}

Each item corresponds to the capability improvement of different sub-modules of the system, and the recovery mechanism that can be adapted and generalized across fault types is realized through multi-objective balance optimization \cite{liu2024mt2st}.

\begin{figure}[h!]
  \centering
  \begin{subfigure}[T]{1\linewidth}
    \includegraphics[width=1\linewidth, height=0.5\linewidth]{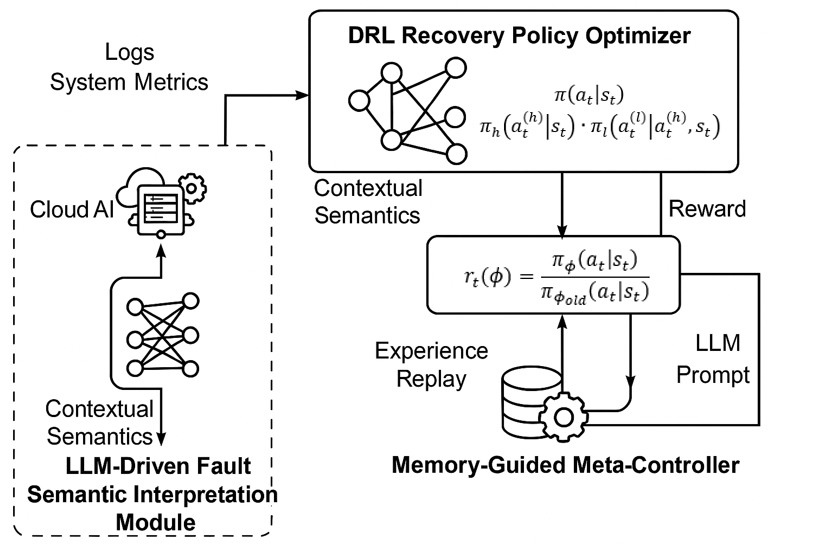}
  \end{subfigure}
  \caption{Intelligent Fault Self-Healing Mechanism (IFSHM) Architecture}
  \label{fig1}
\end{figure}

Figure \ref{fig1} illustrates the overall architecture of an Intelligent Fault Self-Healing Mechanism (IFSHM) for cloud-based AI systems, which integrates large language models (LLMs) and deep reinforcement learning (DRL)\cite{zhao2025bandit}. The LLM-Driven Fault Semantic Interpretation Module on the left side of Figure \ref{fig1} is responsible for extracting contextual semantic information from logs and system monitoring metrics to form fault status vectors. 

The Memory-Guided Meta-Controller below enhances policy generalization capabilities and provides adaptability to new types of faults through empirical playback and prompt tuning mechanisms.

\section{EXPERIMENTS}
\subsection{Experimental setup}

The dataset "Failure-Dataset-OpenStack" provided by the DessertLab team was used in the experiment, which is derived from the fault injection experiment of the OpenStack cloud computing platform, covering a variety of fault types, such as instance faults, network faults, and storage faults, which has high real-world application value. The dataset includes a sequence of fault events and annotated fault information, which reflects the failure behavior of the cloud computing platform in different workload scenarios.

The training and evaluation process is implemented using PyTorch with CUDA 11.7 support on an NVIDIA A100 GPU cluster. The DRL component uses Proximal Policy Optimization (PPO) with a clip ratio of 0.2, learning rate of 3e-4, and batch size of 128. The LLM semantic encoder is based on a 7B-parameter pre-trained model with low-rank adaptation for parameter-efficient fine-tuning. The meta-controller uses a replay buffer of 50K trajectories and applies gradient-based prompt tuning with a temperature of 0.7 and context window size of 1024. Each experiment is repeated five times with random seeds to ensure statistical significance, and average results are reported. Recovery evaluation is measured over 20 injected fault scenarios with varying severity and propagation speed.

In order to comprehensively verify the effectiveness of the proposed intelligent fault self-healing mechanism, we selected the following four models as comparison methods:
\begin{itemize}[leftmargin=1.5em]

\item Deformable DETR (Deformable Detection Transformer) processes time series data and monitoring logs from the cloud platform and uses the self-attention mechanism to accurately capture abnormal patterns in the system.

\item Graph Convolutional Networks for Fault Recovery (GCN-FR) builds a dependency graph of the cloud platform, uses graph convolution to model failure modes, and learns optimal recovery strategies.

\item Transfer Learning for Fault Detection and Recovery (TL-FD/FR) Transfer learning uses transfer learning to migrate fault detection and recovery policies trained on one domain (such as other cloud platforms) to a new environment to reduce the need for large amounts of annotated data.

\item Self-Supervised Learning for Anomaly Detection (SSL-AD) leverages a self-supervised learning framework to automatically generate representations of failure modes by training on large amounts of unlabeled system monitoring data.

\end{itemize}

\subsection{Experimental analysis}
\begin{figure}[h!]
  \centering
  \begin{subfigure}[T]{1\linewidth}
    \includegraphics[width=1\linewidth, height=0.5\linewidth]{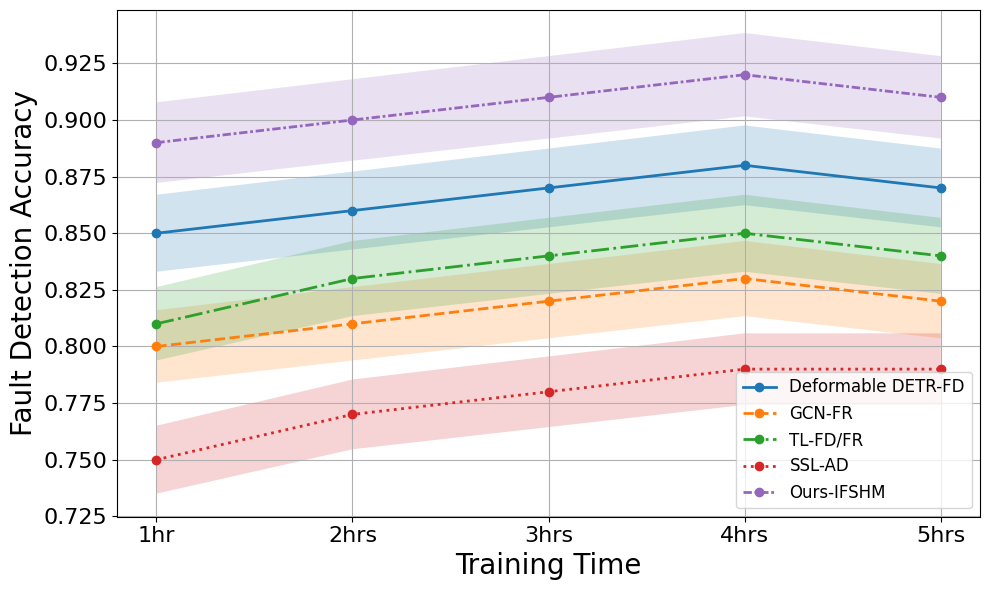}
  \end{subfigure}
  \caption{Fault Detection Accuracy Comparison}
  \label{fig2}
\end{figure}

Fault detection accuracy is a measure of the model's ability to correctly identify faults versus healthy states on a given data set. As can be seen from Figure \ref{fig2}, Ours-IFSHM has always maintained the highest fault detection accuracy at each training duration setting, and the accuracy has steadily improved from 0.89 to 0.92 with the increase of training duration, showing excellent learning and generalization capabilities, while Deformable DETR-FD, GCN-FR, TL-FD/FR, and SSL-AD have also improved at longer training durations. 

However, the accuracy increase is significantly smaller than that of Ours-IFSHM, and the error interval is wider, indicating that they are less robust and stable to parameter changes than the proposed method.

\begin{figure}[h!]
  \centering
  \begin{subfigure}[T]{1\linewidth}
    \includegraphics[width=1\linewidth, height=0.5\linewidth]{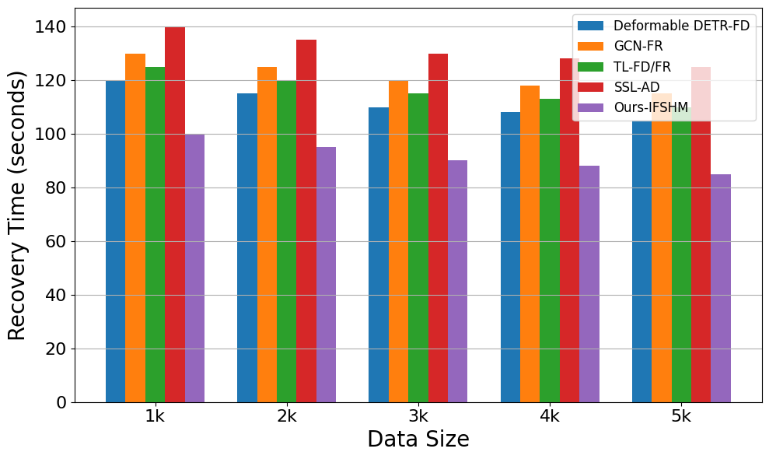}
  \end{subfigure}
  \caption{Recovery Time Comparison}
  \label{fig3}
\end{figure}

Recovery time refers to the time interval between when a failure occurs and when the system returns to normal operation. As can be seen in Figure \ref{fig3}, as the data size increases from 1k to 5k, the recovery time of each method decreases, indicating that the efficiency of the failure recovery mechanism increases with more training data or longer runtime environments. 

Ours-IFSHM has the shortest recovery time, from 100 s to 85 s, which is always ahead of the other four methods; The recovery time of Deformable DETR-FD, GCN-FR, and TL-FD/FR decreased from 120 s, 130 s, and 125 s to 105 s, 115 s, and 110 s, respectively, followed by performance. SSL-AD was the weakest, with recovery times dropping from 140 s to 125 s, but still higher than other methods.

\section{CONCLUSION}
In conclusion, we propose IFSHM, a unified self-healing framework for cloud AI systems that integrates LLM-driven semantic fault interpretation, DRL-based recovery strategy optimization, and memory-guided meta-control. Experimental results based on real OpenStack fault injection data demonstrate that IFSHM outperforms state-of-the-art approaches in both detection accuracy and recovery efficiency. Despite these promising results, this study has some limitations. First, the LLM component is currently fine-tuned offline, which may limit its adaptability to emerging failure types in real-time deployment. Second, the DRL agent requires a substantial number of interaction episodes to generalize well, which can be costly in high-availability systems \cite{ji2025bias}. Additionally, the current architecture assumes centralized log access, which might not hold in federated or privacy-preserving environments. Future research will focus on integrating online continual learning for LLMs, investigating model compression techniques, such as quantization for distributed inference \cite{liu2024contemporary, liu2025llmeasyquant}, for deployment on edge-cloud hybrid environments, and extending the architecture to federated multi-cloud self-healing settings, with a focus on privacy protection mechanisms \cite{luo2025cross}. Furthermore, the optimization of our multi-task objective could be enhanced by exploring more advanced algorithms inspired by recent progress in non-convex optimization \cite{xu2024stochastic, zhang2024agda+}. Exploring zero-shot recovery capabilities and adaptive policy distillation will also be prioritized to further improve robustness and real-time responsiveness. For future optimization of our LLM framework, the multi-agent mechanism could also be introduced to further increase efficiency and precision \cite{chan2023chateval, talebirad2023multi, deng2024composerx, xu2024critique, zhao2024towards, luo2025faithfulpersona}.

\renewcommand{\bibfont}{\footnotesize}

\footnotesize{
\bibliographystyle{IEEEtran}
\bibliography{main}
}

\end{document}